\documentclass[10pt,twocolumn,letterpaper]{article}

\usepackage{cvpr}              %

\definecolor{cvprblue}{rgb}{0.21,0.49,0.74}
\usepackage[pagebackref,breaklinks,colorlinks,allcolors=cvprblue]{hyperref}

\usepackage{enumitem}
\usepackage{booktabs}
\usepackage{caption} %
\usepackage{multirow} %
\usepackage{enumitem}
\usepackage{colortbl} %
\usepackage{bbm}
\usepackage{algorithm}
\usepackage{algpseudocode}   
\usepackage{makecell}
\usepackage{xcolor}

\usepackage{mathtools} %
\usepackage{bm} %
\usepackage{soul} %
\usepackage{nicefrac}
\usepackage{csquotes}

\usepackage{duckuments}
\usepackage{wrapfig}

\usepackage{multirow}
\usepackage{colortbl} %

\def\paperID{22750} %
\def\confName{CVPR}
\def\confYear{2026}

\renewcommand{\thefootnote}{%
  \ifnum\value{footnote}=0
    *
  \else
    \arabic{footnote}
  \fi
}

\title{Which Layer Causes Distribution Deviation? Entropy-Guided Adaptive Pruning for Diffusion and Flow Models}

\author{Changlin Li$^1$$^\ast${\quad}Jiawei Zhang$^2${\quad}Zeyi Shi$^3${\quad}Zongxin Yang$^4${\quad}Zhihui Li$^5${\quad}Xiaojun Chang$^5$$^\ast$\\
{\small
$^1$Stanford University\,\,\,\,\,\,$^2$North China Electric Power University\,\,\,\,\,\,
$^3$University of Technology Sydney}\\{\small$^4$Harvard University\,\,\,\,\,\,$^5$University of Science and Technology of China}\\
}

\begin{document}

\twocolumn[{ 
\maketitle
    \centering
    \includegraphics[width=0.92\linewidth]{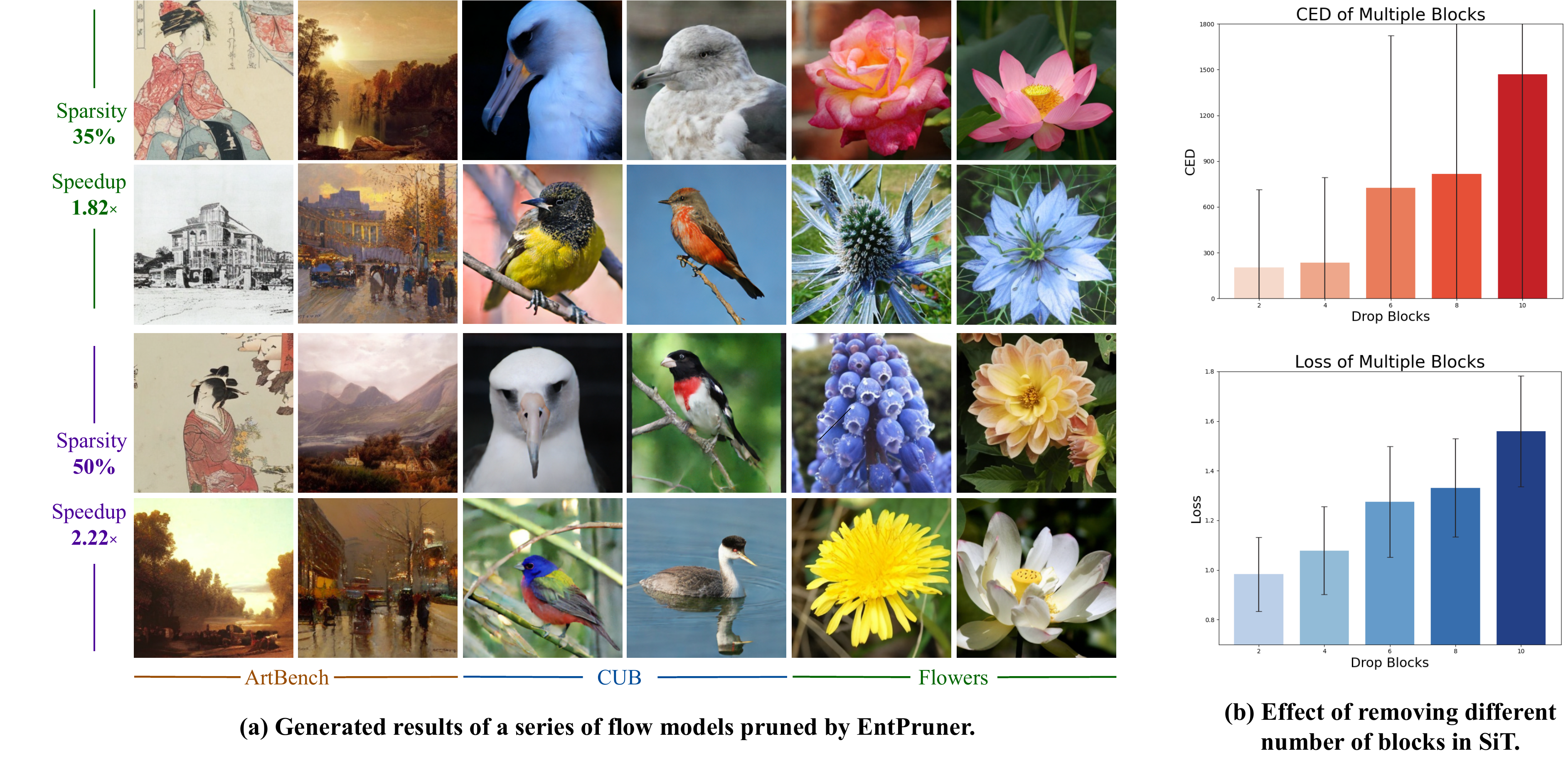}
    \vspace{-15pt}
    \captionof{figure}{
    \textbf{(a) Generated results of a series of  \textit{flow models} pruned by EntPruner.} Entpruner achieves up to 2.22× speedup while maintaining generation quality. Base model: SiT-XL/2 with ODE solver, CFG=4.0, 250 sampling steps. \textbf{(b) Effect of removing different number of blocks in SiT.} Strong correlation between \textit{Conditional Entropy Deviation} (CED) and loss confirms CED's effectiveness in quantifying block importance in flow models.}
    \label{fig:cover_analysis}
\vspace{20pt}
}]

\footnotetext[0]{Correspondance to {\scriptsize\textit{changlinli.ai@gmail.com}}; {\scriptsize\textit{cxj273@gmail.com}}. \\\textbf{Project page:} {\scriptsize\url{https://github.com/changlin31/EntPruner}}}

\begin{abstract}
\vspace{-15pt}

\noindent
Large-scale vision generative models, including diffusion and flow models, have demonstrated remarkable performance in visual generation tasks. However, transferring these pre-trained models to downstream tasks often results in significant parameter redundancy. In this paper, we propose EntPruner, an entropy-guided automatic progressive pruning framework for diffusion and flow models. First, we introduce entropy-guided pruning, a block-level importance assessment strategy specifically designed for generative models.
Unlike discriminative models, generative models require preserving the diversity and condition-fidelity of the output distribution. 
As the importance of each module can vary significantly across downstream tasks, EntPruner prioritizes pruning of less important blocks using data-dependent Conditional Entropy Deviation (CED) as a guiding metric. CED quantifies how much the distribution diverges from the learned conditional data distribution after removing a block.
Second, we propose a zero-shot adaptive pruning framework to automatically determine when and how much to prune during training. This dynamic strategy avoids the pitfalls of one-shot pruning, mitigating mode collapse, and preserving model performance. Extensive experiments on DiT and SiT models demonstrate the effectiveness of EntPruner, achieving up to 2.22× inference speedup while maintaining competitive generation quality on ImageNet and three downstream datasets.

\end{abstract}
    
\vspace{-15pt}
\section{Introduction}
\label{sec:intro}

A myriad of recent breakthroughs in diffusion and flow models have demonstrated their remarkable capabilities in image generation. The success has also been extended to audio, video, and language domains~\citep{huang2023make, zhu2024champ, sahoo2024simple}.
The Denoising Diffusion Probabilistic Model (DDPM)~\citep{ho2020denoising} highlighted the effectiveness of the U-Net backbone. Recently, owing to the superior performance of Diffusion Transformers~\citep{peebles2023scalable},  studies have increasingly adopted transformer-based architectures. Lipman et al.~\citep{lipman2022flow} further introduced Flow Matching, a more direct and faster generative trajectory that offers an alternative perspective on training and inference for diffusion models. While diffusion models have evolved rapidly in recent years, achieving near-photorealistic quality, their practical deployment remains limited due to computational inefficiency. Recent trends in architectural design---particularly the adoption of transformer-based backbones like DiT~\citep{peebles2023scalable} and SiT~\citep{ma2024sit}---have significantly improved scalability and expressivity. However, these advancements come at the cost of increased parameter counts and memory usage. These models suffer from efficiency issues when deployed on edge devices or used in low-latency settings such as interactive applications. The high computational cost and slow inference speed of diffusion and flow models motivate us to explore more effective solutions.

\begin{figure}[t]
    \centering
    \includegraphics[width=.85\linewidth]{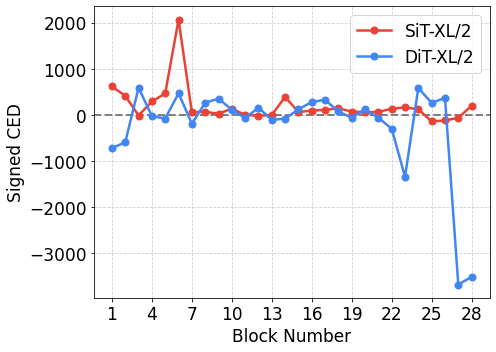}~~~~~~~~~~~~~~~~
    \vspace{-0.5em}
    \caption{\textbf{Signed CED for each block in SiT-XL/2 and DiT-XL/2.} The sign of CED reveals the type of distributional degradation: positive values indicate drift toward randomness/noise (increased entropy), while negative values indicate mode collapse or oversimplified solutions (decreased entropy). Critical blocks exhibit large absolute CED.}
    \label{fig:signed_ced}
    \vspace{-0.15in}
\end{figure}

To solve these problems, a majority of the arts have explored efficient pruning strategies for Stable Diffusion (SD) models~\citep{rombach2022high}. BK-SDM~\citep{kim2024bk} employs a heuristically handcrafted pruning scheme and leverages distillation to recover performance after pruning. However, its manual design limits transferability and requires substantial human efforts and computational costs. Diff-Pruning~\citep{fang2023depgraph} removes filters by identifying unimportant weights through gradient analysis, but its threshold must be tuned for specific tasks, restricting practical applicability. LD-Pruner~\citep{castells2024ld} introduces a novel metric to evaluate the importance of each operator and prunes unimportant convolutional and attention layers, but this metric is task-independent.
Critically, these approaches treat diffusion models as discriminative networks, failing to account for how pruning affects the generative output distribution.
Their reliance on weight magnitudes or gradient-based metrics overlooks the distributional properties that define generative model quality.

These discrepancies raise a critical challenge: \textit{how to adaptively compress diffusion models while preserving their \textbf{distributional expressiveness, diversity, and condition-fidelity}?} Current pruning methods fail to address this in a task-aware and scalable manner, especially when applied to transformer-based diffusion backbones. Furthermore, prior pruning work has also demonstrated a delicate trade-off between model performance and pruning ratio: aggressive pruning often leads to instability and catastrophic forgetting or distribution collapse, making the post-pruning training process slow and ineffective. These challenges highlight the need for a principled, efficient, and robust pruning framework that can adapt to various downstream settings while preserving the learned generative distribution.

Numerous studies have demonstrated that not all parameters in deep neural networks contribute equally to model performance~\citep{cheng2023survey, guo2020model, yanglet}. We observe that removing different blocks causes different types of distribution deviation in generative models. As shown in \cref{fig:signed_ced}, the signed entropy change reveals \textbf{\textit{two distinct patterns}}: \textit{positive values indicate \textbf{drift toward randomness or noise} (increased entropy)}, while \textit{negative values indicate \textbf{mode collapse or oversimplified solutions} (decreased entropy)}.
For instance, SiT-XL/2 exhibits critical positive deviations in mid-layers (block 6), while DiT-XL/2 shows substantial negative deviations in final layers (blocks 27-28), suggesting architectural differences in how blocks maintain distributional balance. This observation motivates us to measure the absolute change of entropy as a generative-specific importance metric that quantifies the magnitude of distribution deviation after removing each block.

We propose a task-aware pruning framework for diffusion and flow models. The framework performs block-level structured pruning through a two-stage process: 1) Entropy-Guided Importance Estimation: We evaluate block importance in the pretrained generative model using Conditional Entropy Deviation (CED), which quantifies how much the distribution diverges from the learned conditional data distribution after removing each block. This enables us to rank blocks based on their contribution to maintaining distributional expressiveness, diversity, and condition-fidelity. 2) Zero-shot Adaptive Pruning: We propose a zero-shot adaptive pruning framework that automatically determines when and how much to prune during training. Unlike one-shot pruning that removes layers at initialization, our progressive approach adapts the pruning schedule dynamically based on training dynamics. To enable efficient pruning decisions without additional validation overhead, we integrate multiple zero-shot metrics, allowing us to reformulate pruning as a lightweight architecture search problem that can be solved with minimal computational cost.

Our framework accurately identifies the least important components for each downstream task, thereby minimizing distribution deviation from the pretrained model. It autonomously determines the optimal pruning schedule at different training stages, resulting in faster convergence while maintaining near-lossless performance. Through extensive experiments on multiple benchmark datasets and two widely-used diffusion architectures (DiT and SiT), we show that our entropy-guided pruning framework not only reduces model size and inference cost but also achieves superior or comparable generative performance compared to full fine-tuning and existing pruning baselines. As demonstrated in \cref{fig:cover_analysis}(a), EntPruner maintains high visual quality even at aggressive pruning ratios, validating that CED-guided adaptive pruning effectively preserves the learned distribution. This work bridges a gap in the compression of diffusion Transformers and offers a practical solution for their deployment in resource-constrained environments. The main contributions are summarized as follows:
\begin{itemize}[leftmargin=*,itemsep=0mm]

\item We introduce \textbf{Conditional Entropy Deviation (CED)}, a generative-specific metric that measures distribution deviation after removing each block, enabling importance ranking that preserves diversity and condition-fidelity during pruning.
\item We propose \textbf{EntPruner}, a zero-shot adaptive pruning framework that automatically determines when and how much to prune during training, achieving stable convergence with minimal loss of distributional quality.

\item Our method achieves an average FID drop of only 1.76 at 50\% pruning rate with \textbf{2.22}$\times$ inference speedup, demonstrating effectiveness across different diffusion and flow models on multiple benchmark datasets.

\end{itemize}

\section{Related Work}\label{sec:related_work}
\noindent\textbf{Diffusion Models}~\citep{ho2020denoising, zhang2023adding, ruiz2023dreambooth} have achieved remarkable progress in image synthesis, with recent designs shifting from U-Nets to Transformer backbones for better scalability~\citep{peebles2023scalable,ma2024sit,chen2024pixartalpha}. While prior works such as DiT, SiT, and PixArt-$\alpha$ have focused on improving generative performance in large-scale diffusion models, our work instead targets the compression and efficient deployment of these Transformer-based architectures through structured Entropy-guided pruning.

\noindent\textbf{Automated Machine Learning. }AutoML automates model design and optimization~\citep{liu2018progressive, tan2019mnasnet, cubuk2019autoaugment}. These techniques collectively reduce the need for manual intervention and enable the efficient discovery of high-performing models across diverse tasks. By simplifying the bi-level optimization problem inherent in AutoML, NAS approaches fall into multi-shot~\citep{real2019regularized}, one-shot~\citep{li2020block,li2021bossnas,peng2021pi}, and zero-shot~\citep{lin2021zen,huang2022arch, yang2024etas, li2024zero} categories. These proxies are often designed based on theoretical insights or structural analysis of deep neural networks, offering a highly efficient alternative to conventional NAS approaches\cite{li2024zero}. In contrast to traditional NAS efforts that aim to discover performant architectures from scratch, our method leverages zero-shot NAS metrics to select and prune subnetworks from pretrained diffusion Transformers, combining efficiency from AutoML with the stability of pretrained models.

\noindent\textbf{Efficient Inference for Diffusion Models. }The inference cost of diffusion models is primarily influenced by the number of inference steps and the computational cost. Existing researches can be broadly categorized into two directions: \textbf{1)} Reducing the number of sampling steps: DPM-Solver\cite{lu2022dpm} advances the numerical approximation of diffusion ODEs. Additionally, \cite{lin2024sdxl, ren2024hyper, li2023snapfusion} employed improved distillation strategy to compress the sampling steps with pre-trained models. \cite{zhang2025blockdance, ma2024learning, liu2025cachequant} introduces a caching mechanism to reuse highly similar features between different sampling steps and reduce redundant computations. \textbf{2)} Compressing model architectures: works\cite{castells2024ld, kim2024bk, fang2023depgraph} applied structured pruning to reduce model size and works\cite{li2023q} adopted quantization techniques to lower both computation and memory requirements. In contrast, \cite{chen2023otov3} proposes a one-shot compression approach that leverages automated pruning and erasure mechanisms to directly extract high-performing sub-networks from any given pre-trained model, without requiring additional fine-tuning. \cite{guo2020model} proposes a Progressive Channel Pruning (PCP) framework that iteratively performs a try-select-prune strategy across multiple layers to gradually compress the network. Our work extends this line by reframing block-level pruning as a zero-shot NAS problem, achieving lightweight yet performant diffusion Transformers.

\section{Entropy-Guided Progressive Pruning for Diffusion and Flow Models}

\subsection{Diffusion and Flow Models}
Both diffusion and flow models corrupt data $\bm{x}_* \sim p(\bm{x})$ by progressively injecting noise $\epsilon \sim \mathcal{N}(0, I)$. A unified forward process can be written as $\bm{x}_t = \alpha_t\bm{x}_* + \sigma_t\epsilon$, where $\alpha_t$ and $\sigma_t$ are time-dependent functions.
In score-matching diffusion models, the process is usually defined on discrete time steps, and $x_t$ converges to $\mathcal{N}(0, I)$ as $t \rightarrow \infty$. Flow-matching methods, in contrast, restrict the trajectory to $t \in [0,1]$ and set $\alpha_0 = \sigma_1 = 1$ and $\alpha_1 = \sigma_0 = 0$, so that $\bm{x}_t$ converges to $\mathcal{N}(0, I)$ precisely at $t = 1$. In the reverse process, both flow matching and score-based diffusion models can be implemented by learning to invert the forward dynamics using either a stochastic differential equation (SDE)~\citep{albergo2022building} or probability flow ordinary differential equation (ODE)~\citep{albergo2023stochastic}.

\begin{figure*}[t]
    \centering
    \includegraphics[width=.95\linewidth]{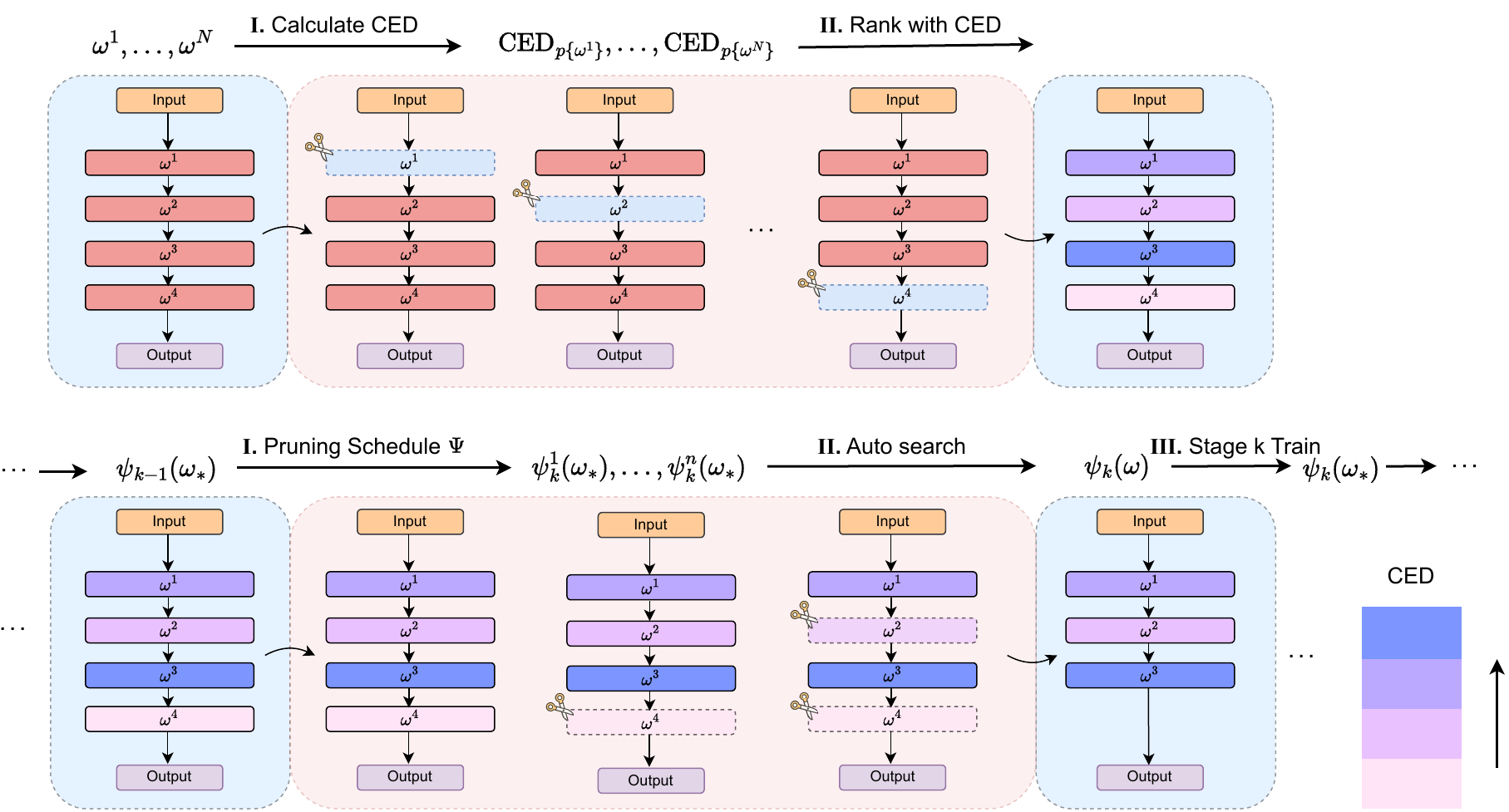}
    \caption{\textbf{Entropy-guided automatic pruning framework. }At first, we employ CED to evaluate and rank the expressiveness of individual blocks, where darker regions in the heatmap indicate stronger interactions with the overall network. At each pruning stage, the pruning schedule explores candidate subnetworks under different pruning ratios, selects the optimal one using zero-shot  proxies, and inherits parameters from the previous stage.}
    \label{fig:entpruner}
    \vspace{-0.15in}
\end{figure*}

\subsection{Overview}
Our goal is to enhance the efficiency of diffusion and flow models by removing blocks that cause minimal distribution deviation, thereby achieving model compression without sacrificing generative quality. To this end, we establish a pruning priority using Conditional Entropy Deviation (CED): given a target pruning ratio $r$, blocks with low CED (minimal distribution deviation when removed) should be pruned first, while high-CED blocks (critical for maintaining the learned distribution) are preserved. 

Prior studies have shown that aggressive one-shot pruning often disrupts the pretrained model's learned distribution, leading to irreversible quality degradation or mode collapse. Therefore, we propose a adaptive progressive pruning framework that gradually removes blocks over the training process. We define a pruning schedule $\Psi$ that consists of a sequence of sub-networks with decreasing model sizes. Let $\mathcal{L}$ denote the task loss, $\Omega$ the model size, and $\bm\omega$ the model parameters. The objective of progressive pruning is:
\begin{equation}
\min_{\bm\omega, \Psi} \left\{ \mathcal{L}(\bm\omega, \Psi),\; \Omega \right\}.
\label{eq:eq1}
\end{equation}

To reduce the complexity of optimizing over $\Psi$, we adopt a linear pruning schedule based on the target pruning ratio $r$. The pruning process is divided into $k$ stages (default $k=4$), with each stage consisting of $s = S/k$ training iterations. The schedule is defined as $\Psi = \{\bm\psi_i\}_{i=1}^{k}$, where $\bm\psi_k$ retains $(1 - r)$\% of the original parameters. 

We optimize the pruning schedule $\Psi$ via zero-shot NAS~\citep{chen2023otov2}, where each candidate subnetwork is evaluated without additional training using metrics such as NTK condition numbers and ZiCo scores~\citep{wang2023dna, li2023zico}. These metrics jointly capture convergence properties and gradient stability, enabling $\Psi$ to automatically determine when and how much to prune while preserving model trainability and distributional quality.

\subsection{Quantifying Block Importance with Conditional Entropy Deviation}

\textbf{Why Entropy for Generative Models?} Unlike discriminative models where layer importance can be assessed through accuracy or gradient magnitudes, generative models require measuring how pruning affects the \textit{output distribution}. For diffusion and flow models, the goal is to preserve the learned conditional distribution $p(\bm{x})$ that captures both sample quality and diversity. We leverage entropy as a natural measure of distributional uncertainty.

\noindent\textbf{Entropy Quantification. }Given a denoising network, let the distribution of output be denoted as $p(x)$, $x \in \mathbf{X}$, where $\mathbf{X}$ represents the output from the network. The entropy is expressed as:
\begin{equation}
\mathcal{H}(\mathbf{X}) = -\int p(\bm{x}) \log p(\bm{x}) \, d\bm{x}, \quad \bm{x} \in \mathbf{X}.
\label{eq:entropy}
\end{equation}
For tractability, we assume that $p(\bm{x})$ follows a Gaussian distribution, i.e., $\mathbf{X} \sim \mathcal{N}(\mu, \sigma^2)$. Eq.(~\ref{eq:entropy}) can be written as:
\begin{equation}
\begin{aligned}
\mathcal{H}(\mathbf{X}) & =-\mathbb{E}[\log\mathcal{N}(\mu,\sigma^{2})] \\
 & =-\mathbb{E}[\log[(2\pi\sigma^2)^{-1/2}\exp(-\frac{1}{2\sigma^2}(f-\mu)^2)]] \\
 & =\log(\sigma)+\frac{1}{2}\log(2\pi)+\frac{1}{2}
\end{aligned}
\end{equation}

\noindent\textbf{Measuring Distribution Deviation with CED. }To quantify how much removing a block disrupts the learned distribution, we introduce \textit{Conditional Entropy Deviation} (CED). CED measures the absolute change in entropy after removing each block, capturing the magnitude of distribution deviation regardless of direction:
\begin{equation}
\begin{aligned}
\mathrm{CED}_i & = |\mathcal{H}_{\text{original}}(\mathbf{X}) - \mathcal{H}_{\text{pruned}}(\mathbf{X}, i)| \\
    & = |\mathcal{H}(\mathbf{X}) - \mathcal{H}(\mathbf{X} \mid \mathtt{Drop}\{\textit{block}_i\})|,
\label{eq:te_definition}
\end{aligned}
\end{equation}
where $\mathcal{H}(\mathbf{X})$ denotes the entropy of the original network output, and $\mathcal{H}(\mathbf{X} \mid \mathtt{Drop}\{\textit{block}_i\})$ represents the entropy after dropping the $i$-{th} block. 

\noindent\textbf{Interpreting CED. } The sign of the entropy change reveals the type of distribution deviation: positive values ($\mathcal{H}_{\text{pruned}} > \mathcal{H}_{\text{original}}$) indicate drift toward randomness or noise, while negative values ($\mathcal{H}_{\text{pruned}} < \mathcal{H}_{\text{original}}$) indicate mode collapse or oversimplified solutions. Both directions signal that the block is critical. For importance ranking, we use the absolute value $\mathrm{CED}_i$: blocks with low CED exhibit minimal distribution deviation when removed (redundant), while high-CED blocks are structurally critical for maintaining the learned distribution.

\subsection{Automatic Pruning via Zero-Shot Metrics. }
Assume that $\mathcal{K}$ is a zero-shot performance predictor used to estimate the loss of each candidate sub-network. The optimal sub-network at stage $k$ can be identified by solving:
\begin{equation}
\begin{aligned}
\bm\psi_k^* & = \mathop{\arg\min}_{\bm\psi_k \in \Lambda_k} \left\{ \mathcal{K}(\bm\omega(\bm\psi_k)),\; \Omega \right\}, \\
\text{where} \quad \Lambda_k & = \left\{ \bm\psi \in \xi \;\middle|\; \left| \bm\omega(\bm\psi) \right| \leq \left| \bm\omega(\bm\psi_{k-1}) \right| \right\}.
\label{eq:zero-min}
\end{aligned}
\end{equation}
where $\Lambda_k$ denotes the set of candidate sub-networks whose parameter sizes are no larger than that of $\bm\psi_{k-1}$, the sub-network from the previous pruning stage.

\noindent\textbf{Trainability via the NTK Condition Number in Flow Matching.} 
The trainability of a neural network reflects how effectively it can be optimized via gradient descent. While larger models offer more expressivity, this does not guarantee practical trainability. The Neural Tangent Kernel (NTK) provides a useful tool for analyzing convergence in the infinite-width regime~\citep{jacot2018neural, novak2022fast}. 

In flow matching, a denoising network learns the velocity field $\mathbf{v}(t, bm{x}_t)$ that transports noisy samples $\bm{x}_t = \alpha_t \bm{x}_* + \sigma_t \epsilon$ toward data $\bm{x}_*$. The true velocity is $\mathbf{v}(t, \bm{x}_t) = \dot{\alpha}_t \bm{x}_* + \dot{\sigma}_t \epsilon$, and the model $\mathbf{v}_\theta(\bm{x}_t, t)$ is trained to approximate it. For a candidate sub-network with parameters $\bm\omega$, the update of predicted velocity satisfies
\begin{equation}
\Delta \mathbf{v}(\bm{x}_t) = -\eta \, \hat{\Theta}(\bm{x}_t, \bm{x}_t) \nabla \mathbf{v}(\bm{x}_t) \mathcal{L},
\end{equation}
where $\hat{\Theta}$ is the NTK of the velocity predictor.

In the infinite-width limit, the training dynamics are governed by the eigenvalues $\{\lambda_i\}$ of $\hat{\Theta}$. The maximum stable learning rate scales as $\eta\sim2/\lambda_0$, while the convergence of the slowest mode depends on $1/\kappa$, where $\kappa = {\lambda_0}/{\lambda_m}.$
A smaller $\kappa$ indicates faster and more stable convergence. Thus, we adopt the NTK condition number as a zero-shot  metric:
\begin{equation}
\mathcal{K}_\kappa(\bm\psi) = \frac{\lambda_0}{\lambda_m}.
\end{equation}
where $\lambda_0$ and $\lambda_m$ denote the largest and smallest eigenvalues of the NTK, respectively. More details are provided in the Supplementary.

\noindent\textbf{Convergence Rate and Generalization Capacity via Gradient Analysis. }Convergence and expressivity also directly influence the final performance of a neural network. After a number of training steps, a network with a larger absolute mean of gradient and smaller standard deviation of gradient is generally associated with lower training loss and faster convergence. Interestingly, a smaller gradient variance often correlates with a lower maximum eigenvalue $\Theta$ of the NTK, which implies a smoother loss landscape and better generalization performance~\citep{lewkowycz2020large}. 

We adopt ZiCo as one of the zero-shot  metrics, which jointly considers the absolute mean and standard deviation of gradient. The parameters of a candidate sub-network $\bm\omega$ are also inherited from the trained parameters $\bm\omega_*$ of the previous pruning stage. Since the ZiCo metric has been shown to correlate positively with network trainability, we introduce a negative sign to make it positively correlated with loss, allowing it to be minimized during optimization:
\begin{equation}
\mathcal{K}_{\mathrm{ZiCo}}(\bm\psi) = - \sum_{l=1}^{N} 
\log \left( \sum_{\bm\omega \in \bm\omega_l} 
\frac{ \mathbb{E}\left[ \left| \nabla_{\bm\omega} \mathcal{L}^* \right| \right] }
{ \sigma \left( \left| \nabla_{\bm\omega} \mathcal{L}^* \right| \right) }
\right).
\end{equation}
where $N$ denotes the number of layers in the candidate sub-network, $\bm\omega_l$ is the set of parameters in the $l^{\text{th}}$ layer, $\mathcal{L}^*$ is the loss $\mathcal{L}(\bm{x}_{t,i}, \mathbf{v}_{t,i}; \bm\omega)$, $i \in \{1, \ldots, D\}$ and $D$ is the number of training batches, typically set to 2 to balance stability and efficiency.
\vspace{-5pt}

\subsection{Entropy-Guided Adaptive Pruning Framework}
\vspace{-5pt}

We summarize the overall algorithmic workflow of EntPruner as follows. First, we perform block-level importance ranking using CED to evaluate the contribution of each block in the pretrained model. Second, we employ two zero-shot  proxies to guide the pruning schedule at each training stage. The optimization objective, as reformulated from  Eq.~(\ref{eq:zero-min}) is:
\begin{equation}
\bm\psi_k^{\star} = \arg\min_{\bm\psi_k \in \Lambda_k} \left\{ \mathcal{K}_{\kappa}(\bm\psi_k),\; \mathcal{K}_{\text{ZiCo}}(\bm\psi_k),\; \Omega \right\}.
\end{equation}
A key challenge lies in how to jointly optimize the two proxies. We assume both proxies are equally important for maintaining network performance and training efficiency. A common strategy is to apply a voting-based algorithm to select the optimal candidate sub-network. This approach helps mitigate differences in scale between the two metrics. Additionally, we incorporate model' parameters as a regularization term in the selection process. The final optimization is defined as:
\begin{equation}
\begin{aligned}
\bm\psi_k^{\star} & = \arg\min_{\bm\psi_k \in \Lambda_k} R(\bm\psi_k), \\
\mathrm{s.t.} R(\bm\psi_{k}) &= R(\mathcal{K}_{\kappa}(\bm\psi_k)) + R(\mathcal{K}_{\text{ZiCo}}(\bm\psi_k)) + \gamma R(\Omega).
\end{aligned}
\label{eq:zero-shot}
\end{equation}
where $R(\cdot)$ denotes the ranking score (\eg, 1st, 2nd, ..., $R$-th) and $\gamma$ is the regularization factor, set to 0.5. The candidate with the lowest rank in each individual metric receives the smallest score, and the sub-network $\bm\psi_k^{\star}$ with the lowest total rank is selected for the next training stage. Please refer to the Supplementary for the Algorithm of EntPruner.%

\section{Experiment}

\subsection{Implementation Details} \label{sec:Setup}

\begin{figure*}[t]
    \centering
    \includegraphics[width=\linewidth]{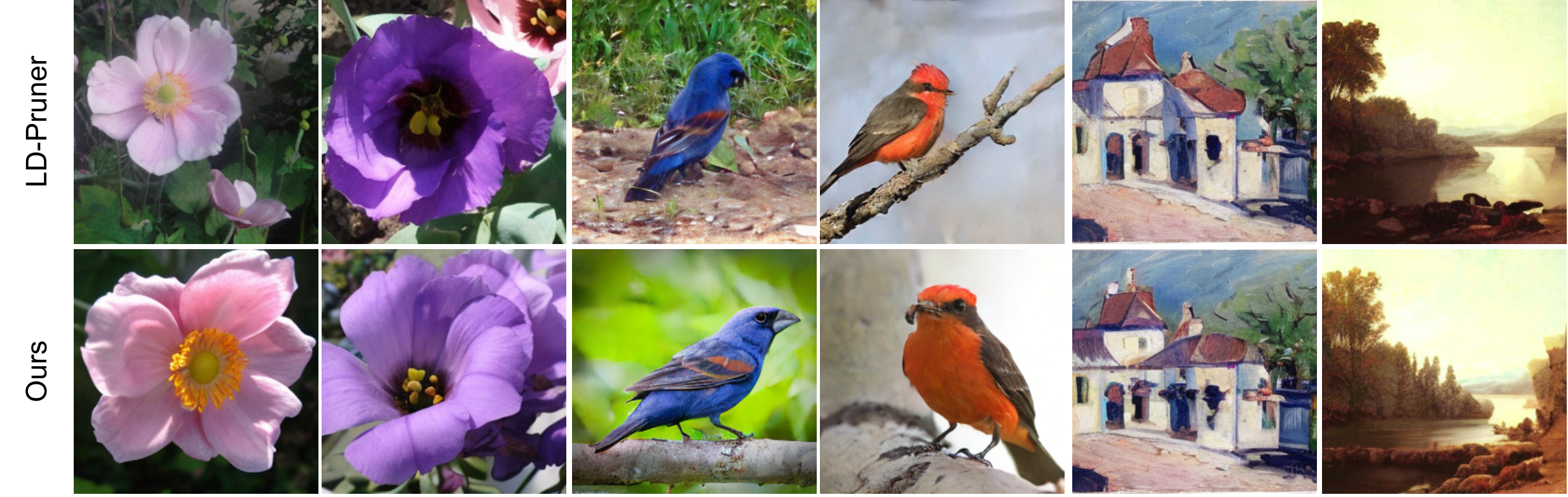}\vspace{-5pt}

    \caption{\textbf{Qualitative comparison of \textit{flow models} pruned by different methods.}  Base model is SiT-XL/2, with 35\% pruning rate. Datasets are Flowers (column 1-2), CUB (column 3-4), and ArtBench (column 5-6). Our EntPruner consistently generates finer details.}
    \label{fig:compare}
    \vspace{-0.15in}
\end{figure*}
\begin{table}[t]
\centering\scriptsize
\setlength{\tabcolsep}{2.5pt}
\begin{tabular}{l|c|cc|cc|cc|cc}
\toprule
\multirow{2}{*}{Method} & \multirow{2}{*}{\makecell{Sparsity}} & \multicolumn{2}{c|}{CUB} & \multicolumn{2}{c|}{Flowers} & \multicolumn{2}{c|}{ArtBench} & \multirow{2}{*}{\makecell{Params\\(M)}} & \multirow{2}{*}{Speedup}\\
 &          &   FID$\downarrow$ & IS$\uparrow$ & FID$\downarrow$ & IS$\uparrow$   & FID$\downarrow$ & IS$\uparrow$  &     &         \\
\midrule
\multicolumn{10}{l}{\textit{w/ ODE sampler}} \\
\midrule
\rowcolor{gray!15}
Fine-tuning     & / & 5.32 & 6.02 & 11.78 & 3.71 & 8.80 & 7.32 & 675.12 & ×1 \\
\midrule
LD-Pruner& \multirow{2}{*}{35\%} & 5.70 & 6.03 & 12.02 & 3.75 & 10.78 & 6.63 & 435.78 & ×1.82  \\
\textbf{Ours}    & & \cellcolor{red!15}\textbf{5.48} & \cellcolor{red!15}\textbf{6.07} & \cellcolor{red!15}\textbf{11.75} & \cellcolor{red!15}\textbf{3.82} & \cellcolor{red!15}\textbf{10.03} & \cellcolor{red!15}\textbf{6.88} & \cellcolor{red!15}\textbf{435.78} & \cellcolor{red!15}\textbf{×1.82}  \\
\midrule
LD-Pruner   & \multirow{2}{*}{50\%} & 6.86 & 6.16 & 12.09 & 3.79 & 12.81 & 6.35 & 334.67 & ×2.22  \\
\textbf{Ours}    &  & \cellcolor{yellow!15}\textbf{6.68} & \cellcolor{yellow!15}\textbf{6.18} & \cellcolor{yellow!15}\textbf{11.86} & \cellcolor{yellow!15}\textbf{3.82} & \cellcolor{yellow!15}\textbf{12.65} & \cellcolor{yellow!15}\textbf{6.41} & \cellcolor{yellow!15}\textbf{334.67} & \cellcolor{yellow!15}\textbf{×2.22}  \\
\midrule
\multicolumn{10}{l}{\textit{w/ SDE sampler}} \\
\midrule
\rowcolor{gray!15}
Fine-tuning    & / & 5.17&5.87 & 12.47&3.77 & 13.33&6.56 & 675.12 & ×1 \\
\midrule
LD-Pruner  & \multirow{2}{*}{35\%} & 5.24&6.10 & 12.32&3.74 & 16.16&6.20 & \textbf{435.78} & ×1.49  \\
\textbf{Ours} &  & \cellcolor{red!15}\textbf{5.22}& \cellcolor{red!15}\textbf{6.11} & \cellcolor{red!15}\textbf{12.10}& \cellcolor{red!15}\textbf{3.74} & \cellcolor{red!15}\textbf{15.25}& \cellcolor{red!15}\textbf{6.31} & \cellcolor{red!15}\textbf{435.78}&
\cellcolor{red!15}\textbf{×1.49}  \\
\midrule
LD-Pruner  & \multirow{2}{*}{50\%} & 5.98 &6.19 & 12.92&3.75 & 18.74&5.90 & \textbf{334.67} & ×1.85  \\
\textbf{Ours}   &  & \cellcolor{yellow!15}\textbf{5.83}& \cellcolor{yellow!15}\textbf{6.15} & \cellcolor{yellow!15}\textbf{12.77}& \cellcolor{yellow!15}\textbf{3.77} & \cellcolor{yellow!15}\textbf{18.69}& \cellcolor{yellow!15}\textbf{5.90} & 
\cellcolor{yellow!15}\textbf{334.67}&
\cellcolor{yellow!15}\textbf{×1.85}  \\
\bottomrule
\end{tabular}
\captionof{table}{\textbf{Comparison of \textit{flow models} pruned by different methods.} We use SiT as the base model and evaluate with both ODE and SDE samplers on three datasets. $\downarrow$ and $\uparrow$ indicate whether lower or higher values are better.}
\label{tab:sit_results}
\end{table}

\begin{table}[t]
\setlength{\tabcolsep}{3pt}
\centering\scriptsize
\vspace{-0.05in}

    \begin{tabular}{l|cccccc}
    \toprule
    Method & \makecell{Sparsity} & CUB & Flowers & ArtBench & Params (M) & Speedup\\
    \midrule
    Full Fine-tuning     & / & 5.68 & 21.05 & 25.31 & 673.8 & ×1 \\
    \midrule
    \rowcolor{green!15}
    \textbf{Ours}     & 30\% & 5.50 & \textbf{11.99} & 24.99 & \textbf{471.66} & \textbf{×1.33}  \\
    \midrule
    Adapt-Parallel & -  & 7.73 & 21.24 & 38.43 & 678.08 & -\\
    Adapt-Sequential & -  & 7.00 & 21.36 & 35.04 & 678.08 & -\\
    BitFit & -   & 8.81 & 20.31 & 24.53 & 674.41 & -\\
    VPT-Deep & - & 17.29 & 25.59 & 40.74 & 676.61 & -\\
    LoRA-R8 & -  & 56.03 & 164.13 & 80.99 & 674.94 & -\\
    LoRA-R16 & - & 58.25 & 161.68 & 80.72 & 675.98 & -\\
    DiffFit & -  & \textbf{5.48} & 20.18 & \textbf{20.87} & 674.63 & -\\ 
     \bottomrule
    \end{tabular}
    
\captionof{table}{\textbf{Comparison of \textit{diffusion models} trained by different methods.} We use DiT as the base model and compare on three datasets. Results are measured in FID.}\label{tab:dit_results}
\label{tab:main}
\end{table}

\textbf{Training.} We evaluate the effectiveness of our proposed method on DiT-XL/2 and SiT-XL/2 models with an image resolution of 256$\times$256. We conduct experiments on both diffusion-based DiT and flow-matching-based SiT.
All experiments are conducted on a computing platform equipped with 8 NVIDIA A800 GPUs (80 GB), with DiT trained for 240K steps and SiT for 60K steps. We fine-tune SiT and DiT on downstream tasks following the configuration in \citep{xie2023difffit, ma2024sit}. 

\noindent\textbf{Evaluation.} For the DiT model, we adopt the DDPM sampler, while for SiT, we employ an ODE solver and SDE solver. We report results based on 50 sampling steps. To assess generative quality, we utilize two widely adopted evaluation metrics: Fréchet Inception Distance (FID) and Inception Score (IS). We measure efficiency using inference latency and the number of parameters.

\noindent\textbf{Datasets.} We evaluate our pruning method on ImageNet and three fine-grained image datasets: CUB-200-2011~\citep{wah2011caltech}, Oxford Flowers~\citep{nilsback2008automated}, and ArtBench-10~\citep{liao2022artbench}. Notably, ArtBench-10 exhibits a distribution that significantly differs from ImageNet, allowing us to comprehensively evaluate the generalization performance of EntPruner on out of distribution tasks.

\subsection{Entropy-Guided Adaptive Pruning on Downstream Datasets}

\textbf{Class to Image Generation with Flow Models. }To evaluate the effectiveness of our approach, we compare it with LD-Pruner~\citep{castells2024ld} on the SiT architecture. As shown in \cref{tab:sit_results}, our method consistently outperforms LD-Pruner at both 35\% and 50\% pruning ratios, when pruning 35\% of the parameters, we achieve 3.9\%, 2.2\%, and 6.9\% FID improvements on the three benchmark datasets with an ODE sampler. Notably, on Flowers dataset, our method surpasses full fine-tuning when pruning 35\% of the parameters, achieving the best overall performance. We observe slight variations across different samplers, with the SDE sampler exhibiting longer sampling times compared to the ODE. Nevertheless, our method consistently outperforms LD-Pruner regardless of the sampling strategy. While both pruning methods exhibit performance degradation on ArtBench, likely due to the substantial distribution gap between ArtBench and ImageNet, EntPruner still outperforms LD-Pruner, demonstrating strong robustness and superior generalization. 

We qualitatively compare the image generation quality of EntPruner and LD-Pruner. In these experiments, we use the same random seed, set the classifier-free guidance scale to 4.0, and adopt 250 sampling steps. As shown in ~\cref{fig:compare}, our method consistently produces images with finer details and better quality. On the CUB and Oxford Flowers datasets, EntPruner tends to generate close-up views of the subject, enriching detail and enhancing aesthetic quality. On the ArtBench dataset, our approach captures more realistic textures in structures such as houses and trees, resulting in improved visual fidelity compared to LD-Pruner.

\noindent\textbf{Class to Image Generation with Diffusion Models. }We apply EntPruner on DiT and
compare with full fine-tuning.
As shown in \cref{tab:dit_results}, our method outperforms full fine-tuning across all three datasets. Notably, on Flowers, EntPruner reduces the FID relatively by \textbf{43.04\%}. Furthermore, EntPruner improves inference efficiency, achieving a \textbf{1.33$\times$} speedup compared to full fine-tuning. 
In addition, we benchmark EntPruner against \textit{state-of-the-art} fine-tuning strategies. Our method matches or surpasses the performance of efficient fine-tuning baselines, while offering \textbf{1.33$\times$} faster inference.

\begin{figure}[t]
    \centering
\vspace{-5pt}
\includegraphics[width=0.95\linewidth]{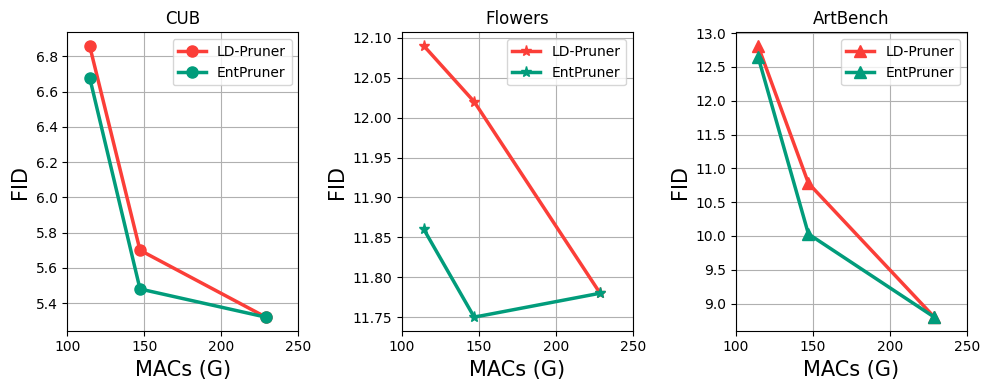}\vspace{-8pt}
    \caption{\textbf{Tradeoff between computational cost (MACs) and generative quality (FID).} Rightmost point: full model.}    
    \vspace{-5pt}

    \label{fig:fid_macs}
\end{figure}

\noindent\textbf{Inference Efficiency Across Parameter Budgets. }\cref{fig:fid_macs} compares the FID scores \textit{vs.} computational complexity (measured by MACs) of EntPruner and LD-Pruner across three datasets: CUB, Flowers, and ArtBench. Notably, EntPruner consistently outperforms LD-Pruner across all configurations. At~medium complexity, the gap between the two is most obvious. \cref{tab:sit_results} and \cref{fig:fid_macs} jointly demonstrate that EntPruner maintains lower FID scores even under reduced MACs on CUB and Flowers datasets, demonstrating better robustness in resource-constrained settings. On ArtBench dataset, the performance gap between the two methods is most pronounced, further highlighting EntPruner’s superior generalization ability and pruning efficiency for complex image generation tasks. See the Supplementary for more details.

\vspace{-5pt}
\subsection{Entropy-Guided Adaptive Pruning on ImageNet 256×256}\vspace{-5pt}
\begin{figure*}[t]
    \centering
    \includegraphics[width=\linewidth]{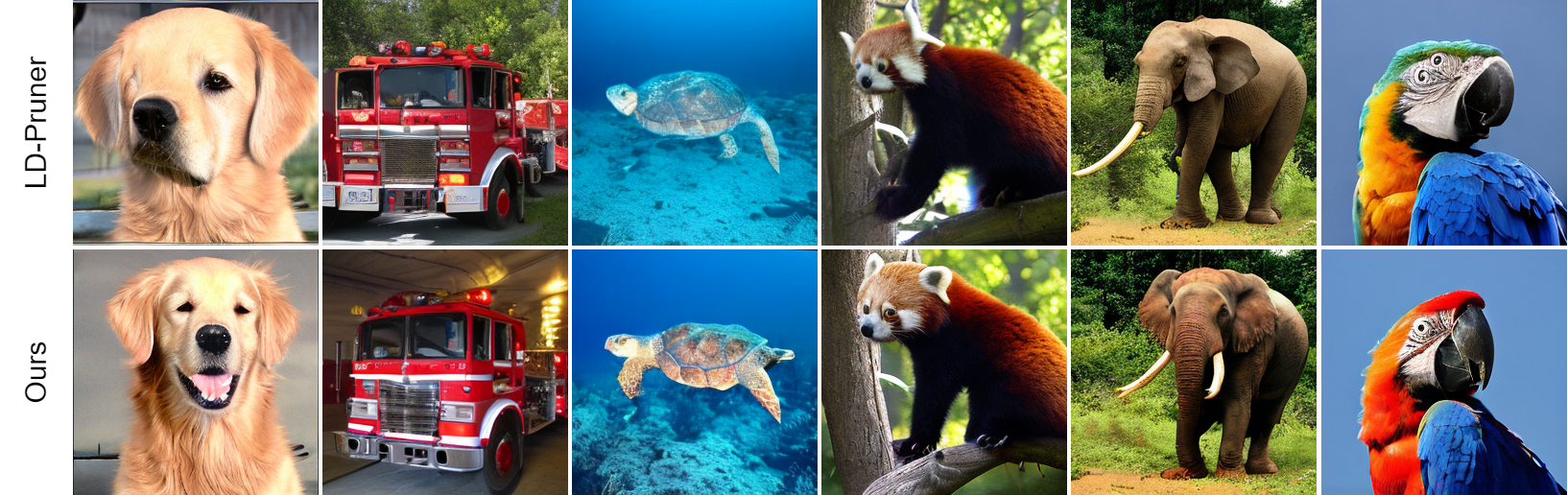}

    \caption{\textbf{Qualitative comparison of \textit{flow models} pruned by different methods on ImageNet 256×256.} Base model is SiT with 30\% pruning rate. Our EntPruner consistently produces more photorealistic images with \textit{fewer generation artifacts}.}
    \label{fig:imagenet_compare}
\end{figure*}

To further demonstrate the effectiveness and robustness of our pruning method on pretrained models, we directly apply our pruning strategy to compress models pretrained on ImageNet 256$\times$256. The experimental setup follows the same configuration as \cref{sec:Setup}, where SiT is trained using flow matching. During sampling, we employ an ODE solver and adhere to standard evaluation protocols~\citep{peebles2023scalable, ma2024sit}.

As shown in ~\cref{tab:sit_imagenet}, with a pruning ratio of 30\% on ImageNet, our method achieves a final FID of \textbf{3.53}, only a degradation of 1.38 compared to the original SiT-XL/2. Notably, it surpasses the recent pruning method LD-Pruner by 48.16\%, further validating our method's ability to mitigate parameter collapse often caused by one-shot pruning. Moreover, in terms of inference speed, our method achieves a 1.33$\times$ speedup compared to SiT, and a \textbf{209.30\%} speed improvement compared to DiT, highlighting its practical inference efficiency.

As shown in ~\cref{fig:imagenet_compare}, we present qualitative results demonstrating the impact of pruning on image generation quality. It is evident that pruning with LD-Pruner leads to significant degradation in visual fidelity. For instance, generated images exhibit missing features such as the eyes of a dog, fine details of a raccoon, aesthetically pleasing text on hot air balloons, well-structured cakes, and the tusks of an elephant. In contrast, our method preserves more structural details and visual coherence. By pruning at more appropriate stages, our approach enables the final model to maintain robust generative performance while retaining essential semantic content.

\subsection{Ablation Study}

\begin{table}[t]
\centering\footnotesize

    \begin{tabular}{l|cccc}
    \toprule
    Method & Params(M) & FID$\downarrow$ & MACs &Speedup\\
    \midrule
    BigGAN-deep & 112 & 6.95 & - & - \\ 
    \midrule
    AR w/ VQGAN & 227 & 26.52 & - & - \\
    MaskGIT & 227 & 6.18 & - & - \\
    \midrule
    ADM & 554 & 10.94  & - & -\\
    LDM-4-G & 400 & 3.60  & - & -\\
    DiT-XL/2 (DDPM)& 675.12 & 2.27  & 228.85 & ×0.43\\
    SiT-XL/2 (ODE) & 675.12 & 2.15  & 228.85 & ×1\\
    \midrule
    LD-Pruner (SiT,ODE) & 471.66 & 6.81  & 163.48 & ×1.33\\
    \rowcolor{green!15}
    Ours (SiT,ODE) & \textbf{471.66} & \textbf{3.53}  & \textbf{163.48} & \textbf{×1.33}\\
     \bottomrule
    \end{tabular}
\caption{
\textbf{Comparison of \textit{flow models} pruned by different methods on ImageNet 256×256.} Base model is SiT with 30\% pruning rate. We also include full model performance of different generative models as a reference.
}\label{tab:sit_imagenet}
\vspace{-10pt}
\end{table}

\textbf{Entropy-Guided CED ranking. }We perform an ablation study to evaluate the effectiveness of entropy-guided importance ranking. Random pruning leads to high variance and instability, so we prune blocks with the highest Conditional Entropy Deviation, which are identified as most critical by our metric. As illustrated in \cref{tab:ablation}, pruning blocks with high entropy causes a substantial performance drop, which cannot be fully recovered even through continued fine-tuning. which proves that Conditional Entropy Deviation is a reliable indicator of parameter importance in pretrained networks.

\begin{table}[t]
\centering
\footnotesize

\setlength{\tabcolsep}{3pt}
    \begin{tabular}{l|cccccc}
    \toprule
    Method & FID $\downarrow$ & IS $\uparrow$ & \makecell{Params\\(M)} & MACs & \makecell{latency\\(s)} & Speedup\\
    \midrule
    Full Fine-tuning     & 11.78 & 3.71 & 675.12 & 228.85 & 0.20 & ×1 \\
    \midrule
    w/o CED     & 12.06 & 3.81 & 435.78 & 147.13 & 0.09 & ×1.82  \\
    w/o Ada. Pruning     & 11.84 & 3.80 & 435.78 & 147.13 & 0.09 & ×1.82 \\
    \midrule
    \rowcolor{green!15}
    Ours     & \textbf{11.75} & \textbf{3.82} & \textbf{435.78} & \textbf{147.13} & \textbf{0.09} & \textbf{×1.82} \\
     \bottomrule
    \end{tabular}
\vspace{-5pt}\caption{
\textbf{Ablation studies on CED ranking and adaptive pruning.} Results are reported with SiT on Flowers. Latency denotes the sampling time for an images executed on a single A800 GPU.%
}\label{tab:ablation}
\vspace{-10pt}
\end{table}

\noindent\textbf{Adaptive Pruning. }We conduct an ablation study on automated pruning. The results are presented in \cref{tab:ablation}, demonstrating that one-shot pruning or premature pruning significantly degrades model performance. In contrast, our method autonomously determines both the timing and extent of pruning, allowing the model to maintain competitive performance, or even achieving better results than full fine-tuning in some cases.

\section{Conclusion}
We present EntPruner, an entropy-guided adaptive pruning framework specifically designed for diffusion and flow models. By introducing Conditional Entropy Deviation (CED), we provide a generative-specific metric that measures distribution deviation after removing each block—capturing both drift toward randomness and mode collapse. Our zero-shot adaptive pruning framework automatically determines when and how much to prune during training, preserving distributional quality, diversity, and condition-fidelity while achieving up to 2.22$\times$ inference speedup. Extensive experiments demonstrate that our method maintains generation quality comparable to full fine-tuning across multiple benchmarks. 

\noindent\textbf{Limitation.} The ease of deployment may enable misuse in unregulated or adversarial scenarios. Future work could explore extending CED to other generative architectures beyond transformer-based diffusion models.

{
    \small
    \bibliographystyle{ieeenat_fullname}
    \bibliography{main}
}

\appendix
\clearpage
\setcounter{page}{1}
\maketitlesupplementary

\begin{figure*}[t]
    \centering
    \includegraphics[width=.95\linewidth]{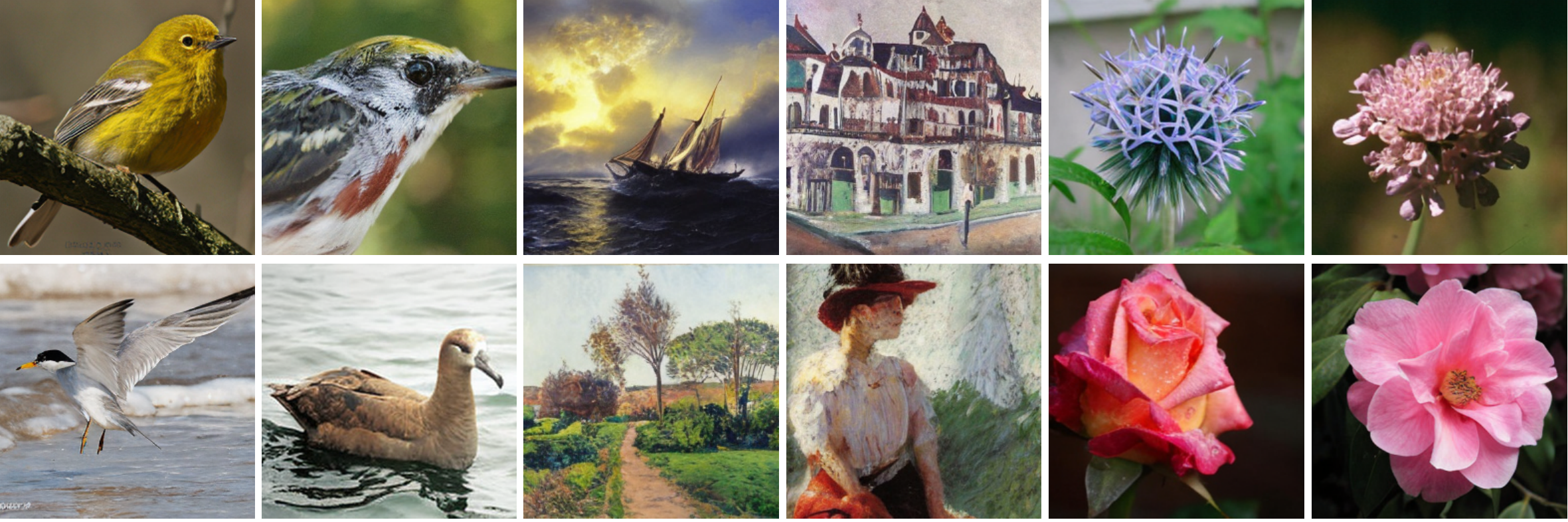}
    \caption{\textbf{Generated results of a series of \textit{diffusion models} pruned by EntPruner.} Base model is DiT-XL/2. The pruning rates are set to 30\%. Inference is performed on the CUB (column 1-2), ArtBench (columns 3-4), and Flowers (columns 5-6)  datasets, with the classifier-free guidance coefficient set to 4.0. The sampling process involves 250 steps.}
    \label{fig:dit_results_images}
\end{figure*}

\section{Appendix}
\label{sec:Appendix}

\subsection{More Qualitative Results}

The sampling results produced by applying our method to the DiT model are shown in Figure \ref{fig:dit_results_images}. On CUB dataset (columns 1–2), the generated results exhibit fine-grained feather textures and natural color transitions. On ArtBench (columns 3–4), our method produces artwork with complete scene structure and stylistic consistency. On Flowers dataset (columns 5–6), the generated results display vibrant and well-separated color tones and clear petal boundaries.

\subsection{Performance of Different Methods}

\begin{figure}[t]
    \centering
    \includegraphics[width=\linewidth]{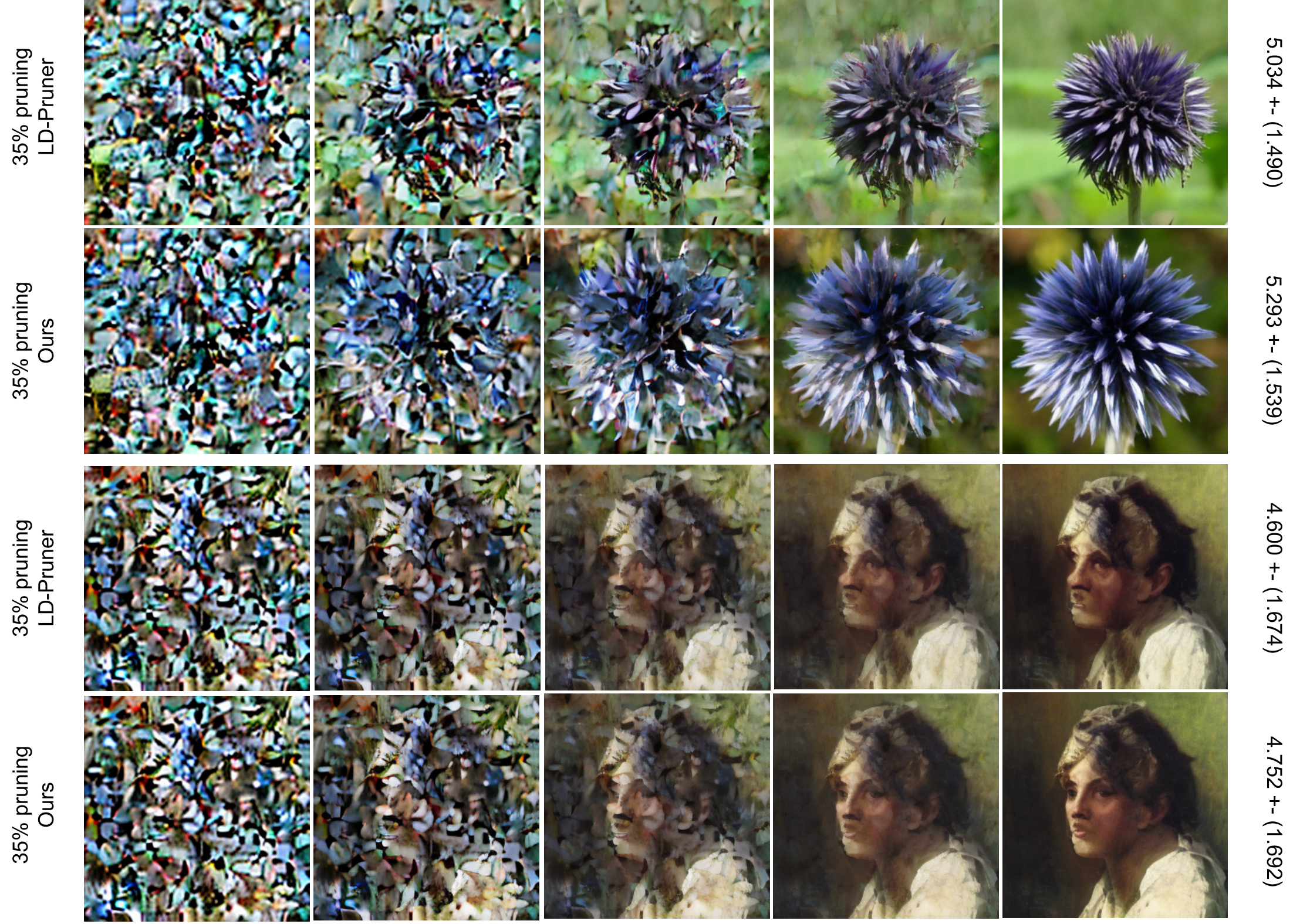}
    \caption{\textbf{Visualization of the sampling process of SiT models pruned with different methods.} From left to right, each column corresponds to sampling steps 50, 100, 150, 200 and 250, respectively.}
    \label{fig:sit_process_gen}
\end{figure}

\begin{figure}[t] 
    \centering
    \includegraphics[width=.99\linewidth]{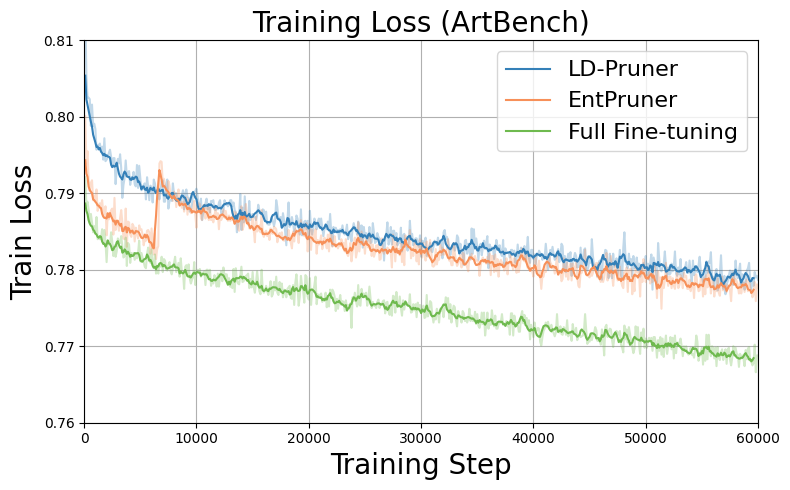}
    \caption{Comparison of training loss trajectories across different methods.}
    \label{fig:loss_curve}
\end{figure}

To better understand the generative behavior of different methods, we visualize the denoising trajectories starting from the same noise latent. As shown in Figure~\ref{fig:sit_process_gen}, we track the intermediate outputs along the denoising path for each method. While both models begin from identical noise, our method consistently produces more visually appealing and coherent results compared to LD-Pruner. In addition, we evaluate the aesthetic quality of generated images using the Neural Image Assessment (NIMA) metric~\citep{talebi2018nima}. NIMA employs a trained deep convolutional neural network to predict how users would rate an image in terms of technical quality and aesthetic appeal. Experimental results show that our method outperforms LD-Pruner by 0.259 and 0.152 points in aesthetic scores, respectively, demonstrating the effectiveness of our approach in preserving visual quality. Figure \ref{fig:loss_curve} illustrates the loss trajectory during training with SiT on ArtBench dataset. Owing to our automated pruning strategy, model compression is performed progressively, avoiding abrupt parameter collapse. This leads to faster and more stable convergence.

\subsection{Inference Efficiency Across Parameter Budgets. }\label{sec:inference_efficiency}

We evaluate the inference efficiency of our method applied to both SiT and DiT under varying parameter scales. FID is reported as the average across three benchmark datasets. Multiply–Accumulate Operations (MACs) are used to quantify the computational complexity of each model. In addition, per-image inference time is measured with a batch size of 256. Peak memory usage is recorded as maximum GPU memory required for a single-image inference on A800.

\begin{table}[t]
    \centering
    \small
    \setlength{\tabcolsep}{3pt}
    \caption{\textbf{Inference Efficiency Evaluation.} We apply our proposed method to both SiT and DiT models to systematically assess inference efficiency under varying parameter budgets.}%
    \begin{tabular}{l|ccccc}
    \toprule
    Method & \makecell{Params} & \makecell{Avarage\\FID} & \makecell{MACs\\(G)} & \makecell{Latency\\(s/per)} & \makecell{Memory\\(GB/per)}\\
    \midrule
    \multirow{2}{*}{DiT/DDPM} &  675.12 & 17.35 & 228.85 & 0.41 & 5.35   \\
                              &  \cellcolor{green!15}483.88 & \cellcolor{green!15}14.16 & \cellcolor{green!15}163.48 & \cellcolor{green!15}0.31 & \cellcolor{green!15}4.56    \\
    \midrule
    \multirow{3}{*}{SiT/ODE}  & 675.12 & 8.63 & 228.85 & 0.20 & 5.32     \\
                              & \cellcolor{red!15}435.78 & \cellcolor{red!15}9.08 & \cellcolor{red!15}147.13 & \cellcolor{red!15}0.11 & \cellcolor{red!15}4.39     \\
                              & \cellcolor{yellow!15}334.67 & \cellcolor{yellow!15}10.39 & \cellcolor{yellow!15}114.45 & \cellcolor{yellow!15}0.09 & \cellcolor{yellow!15}3.99    \\
    \midrule
    \multirow{3}{*}{SiT/SDE} &  675.12 & 10.32 & 228.85 & 0.76 & 5.36    \\
                              &  \cellcolor{red!15}435.78 & \cellcolor{red!15}10.85 & \cellcolor{red!15}147.13 & \cellcolor{red!15}0.51 & \cellcolor{red!15}4.38   \\
                              &  \cellcolor{yellow!15}334.67 & \cellcolor{yellow!15}12.43 & \cellcolor{yellow!15}114.45 & \cellcolor{yellow!15}0.41 & \cellcolor{yellow!15}3.93   \\
     \bottomrule
    \end{tabular}
\label{tab:efficiency_of_inference}
\end{table}

As shown in Table~\ref{tab:efficiency_of_inference}, on DiT, pruning 30\% of parameters leads to an 18.4\% improvement in average FID compared to full fine-tuning. This result reinforces the insight that large-scale models often contain substantial parameter redundancy when transferred to downstream tasks—redundancy that can hinder rather than help model performance. Our pruning strategy is block-level, which allows the computational complexity (as measured by MACs) to decrease proportionally with reduction in parameter count. For instance, when the parameter count is reduced by 35\% and 50\%, the corresponding MACs are also reduced by approximately 35\% and 50\%. We further observe that for SiT, using the Euler-based SDE sampler results in the slowest inference speed, whereas the ODE sampler offers the fastest. At a pruning rate of 35\%, the FID scores of both samplers decrease (by 0.53 and 0.45), while the memory usage decreases by 18.3\% and 19.8\%.

\subsection{Trainability via the Condition Number of NTK in Flow Matching.}\label{sec: NTK}
The trainability of a neural network reflects its ability to be effectively optimized via gradient descent. While a network with more parameters theoretically possesses greater expressivity, this does not guarantee practical trainability. The Neural Tangent Kernel (NTK) provides a powerful analytical tool for assessing the convergence behavior of deep networks under gradient-based optimization, particularly in infinite-width regime~\citep{jacot2018neural, novak2022fast}.

During progressive pruning, we evaluate the trainability of a candidate sub-network with parameters $\omega \in \Lambda_k$, inherited from the previous training stage. Let $\mathcal{L}$ be the velocity prediction loss. Using the chain rule, the parameter update $\Delta \omega$ and the corresponding change in predicted velocity $\Delta v$ can be expressed as:
\begin{equation}
\begin{aligned}
\Delta\boldsymbol{\omega} &= -\eta \nabla_{\boldsymbol{\omega}} v(\boldsymbol{x}_t)^{\mathsf{T}} \nabla{v(\boldsymbol{x}_t)} \mathcal{L}, \\
\Delta v(\boldsymbol{x}_t) &= \nabla{\boldsymbol{\omega}} v(\boldsymbol{x}_t) \Delta \boldsymbol{\omega} \\
&= -\eta \hat{\Theta}(\boldsymbol{x}_t, \boldsymbol{x}_t) \nabla{v(\boldsymbol{x}_t)} \mathcal{L},
\end{aligned}
\end{equation}
where $\eta$ is the learning rate, and $\hat{\Theta}(\boldsymbol{x}_t, \boldsymbol{x}_t) = \nabla_{\boldsymbol{\omega}} v(\boldsymbol{x}_t)^{\mathsf{T}} \nabla_{\boldsymbol{\omega}} v(\boldsymbol{x}_t)$ represents the NTK of the velocity prediction network.

In the infinite-width limit, the NTK remains constant throughout training, and the expected output $\mu(x_t)$ evolves as \cite{jacot2018neural}:
\begin{equation}
\mu(x_t) = (\mathbf{I} - e^{-\eta \hat{\Theta} s}) v_t,
\end{equation}
where $s$ is the training step index. This expression can be diagonalized in the eigenspace of $\hat{\Theta}$, yielding:
\begin{equation}
\mu(x_t)_i = (1 - e^{-\eta \lambda_i s}) v_{t,i},
\end{equation}
where $\lambda_i$ denotes the $i^\text{th}$ eigenvalue of the NTK.

By ordering the eigenvalues as $\lambda_0 \geq \cdots \geq \lambda_m$, it is known that the maximum stable learning rate scales as $\eta \sim 2 / \lambda_0$. Consequently, the convergence rate of the slowest mode is governed by $1 / \kappa$, where $\kappa = \lambda_0 / \lambda_m$ is the condition number of the kernel. A smaller $\kappa$ indicates better trainability and faster convergence.

Thus, we adopt the NTK condition number as one of our zero-shot NAS metrics:
\begin{equation}
\mathcal{H}\_\kappa(\psi) = \frac{\lambda_0}{\lambda_m},
\end{equation}
where $\lambda_0$ and $\lambda_m$ denote the largest and smallest eigenvalues of the NTK, respectively.

\end{document}